\pdfoutput=1

\documentclass[11pt]{article}

\usepackage[final]{acl}

\usepackage{times}
\usepackage{latexsym}
\usepackage{booktabs}
\usepackage{multirow}
\usepackage{amsmath}
\usepackage{url}

\usepackage[T1]{fontenc}

\usepackage[utf8]{inputenc}

\usepackage{microtype}

\usepackage{inconsolata}

\usepackage{graphicx}

\DeclareMathOperator*{\argmax}{arg\,max}

\title{Analysing zero-shot temporal relation extraction on clinical notes using temporal consistency}

\author{
 \textbf{Vasiliki Kougia\textsuperscript{1,2,*}},
 \textbf{Anastasiia Sedova\textsuperscript{1,2}},
 \textbf{Andreas Stephan\textsuperscript{1,2}},
 \\
 \textbf{Klim Zaporojets\textsuperscript{3}},
 \textbf{Benjamin Roth\textsuperscript{1,4}}
 \\
 \\
 \textsuperscript{1}Faculty of Computer Science, University of Vienna, Vienna, Austria
 \\
 \textsuperscript{2}UniVie Doctoral School Computer Science, Vienna, Austria
 \\
 \textsuperscript{3}Department of Computer Science, Aarhus University, Aarhus, Denmark
 \\
 \textsuperscript{4}Faculty of Philological and Cultural Studies, University of Vienna, Vienna, Austria
 \\
 \textsuperscript{*}vasiliki.kougia@univie.ac.at
}

\begin{document}
\maketitle
\begin{abstract}

This paper presents the first study for temporal relation extraction in a zero-shot setting focusing on biomedical text. 
We employ two types of prompts and five LLMs (GPT-3.5, Mixtral, Llama 2, Gemma, and PMC-LLaMA) to obtain responses about the temporal relations between two events. 
Our experiments demonstrate that LLMs struggle in the zero-shot setting performing worse than fine-tuned specialized models in terms of F1 score, showing that this is a challenging task for LLMs.
We further contribute a novel comprehensive temporal analysis by calculating consistency scores for each LLM. Our findings reveal that LLMs face challenges in providing responses consistent to the temporal properties of uniqueness and transitivity. Moreover, we study the relation between the temporal consistency of an LLM and its accuracy and whether the latter can be improved by solving temporal inconsistencies. 
Our analysis shows that even when temporal consistency is achieved, the predictions can remain inaccurate.
\end{abstract}

\begin{figure}
\includegraphics[scale=0.3]{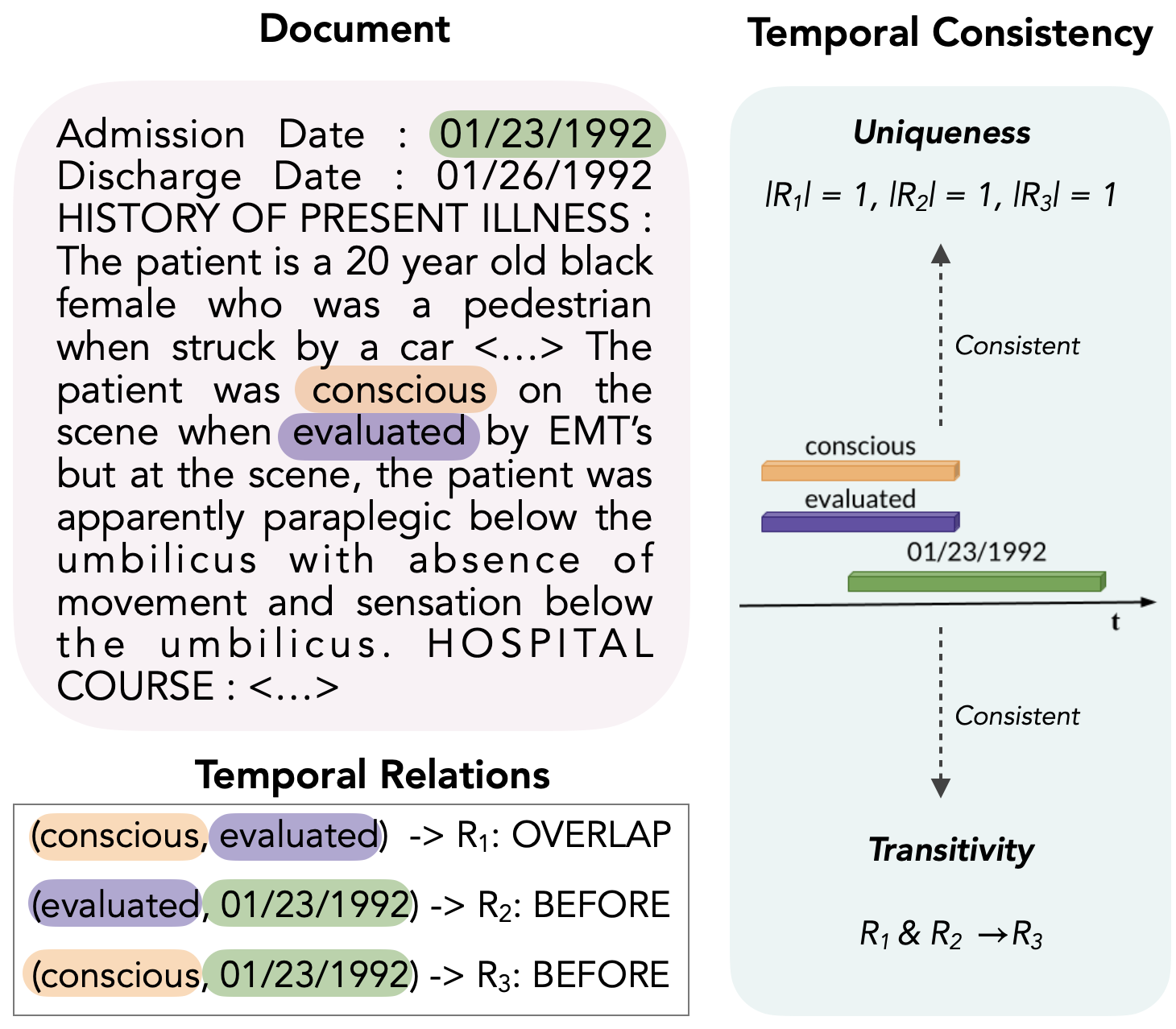}
    \caption{An example of three event pairs annotated with temporal relations. In the right part, the order of the events with respect to time (t) is shown and the consistency of uniqueness and transitivity.}
    \label{fig:first_page}
\end{figure}

\section{Introduction}

Reasoning regarding the temporality of events detected in a text (e.g., understanding their duration, frequency, and order) is an essential part of natural language understanding \cite{Allen1983,Wenzel2023}. Event ordering can be approached as identifying temporal relations between two events, a task often referred to as \textit{temporal relation extraction} (TempRE). 
This task can also be applied to medical text (BioTempRE), e.g., clinical notes written by clinicians regarding a patient's visit, and various medical events such as symptoms, treatments, tests, and other medical terms (see Figure~\ref{fig:first_page}).
BioTempRE has numerous useful applications in healthcare and can assist in medical diagnosis, including adverse drug event detection and medical history construction \cite{Sun2013,Gumiel2021survey,Haq2021,Tu2023}.
Current state-of-the-art methods perform supervised learning, which requires annotated datasets \cite{Wang2022,Yao2022,Knez2024}. However, acquiring high-quality annotated data for TempRE poses significant challenges causing problems to existing datasets like missing relations and low inter-annotator agreement \cite{Ning2017}. In the biomedical domain, this challenge is aggravated by the need for expert knowledge and the sensitive nature of medical data.

In TempRE, there are important properties that emerge from the temporal nature of this task and determine the relations between events (Figure \ref{fig:first_page}). Such properties are \textit{symmetry} (e.g., A \textit{BEFORE} B $\Rightarrow$ B \textit{AFTER} A) and \textit{transitivity} (e.g., A \textit{BEFORE} B and B \textit{BEFORE} C $\Rightarrow$ A \textit{BEFORE} C). We also identify the property of \textit{uniqueness}: each pair of events can have only one temporal relation since they are mutually exclusive. These properties can be utilized to enforce global \textit{temporal consistency} on a model's predictions: for example, on a unified output of different classifiers \cite{Chambers2014,Tang2013}, on a model that operates locally (i.e., with one pair of events as input,  \citet{Ning2017}), or on predicted relations between different types of events \cite{Wang2022}.

Recently, LLMs have shown remarkable performance in several tasks even in a zero-shot setting, which helps to tackle the need for training data \cite{Bubeck2023,wei2022finetuned}.
Numerous works experiment with predictions of LLMs and study their reasoning abilities and the impact of various prompts in different tasks \cite{Wu2023,Jain2023,Tan2023}. Despite the success of LLMs, studies show that these models continue to face challenges in temporal reasoning, especially in TempRE \cite{Tan2023,Jain2023,Yuan2023}, as well as in biomedical tasks \cite{Wu2023}. 
In zero-shot TempRE, \citet{Yuan2023} employed different prompts for ChatGPT and found that it has a considerably lower performance compared to standard supervised methods. They also report ChatGPT's tendency to provide temporally inconsistent responses, but did not perform an evaluation of temporal consistency specifically. Moreover, none of the mentioned papers investigated the temporal reasoning capabilities of LLMs on medical data.

In this paper, we perform zero-shot BioTempRE on clinical notes (i.e., medical texts documenting patients visits) by using prompts consisting of a clinical note and questions regarding which temporal relation exists between a pair of events.\footnote{We do not perform event detection and instead consider the events in each text already known.} We experiment with two different prompting strategies (BatchQA and CoT) and five widely-used LLMs (GPT-3.5, Mixtral 8x7B, Llama2 70B, Gemma 7B, and PMC-LLaMA 13B). Our findings reveal a poor performance of LLMs in this task with a difference of approximately 0.2 in F1 score compared to supervised models. 
Furthermore, we calculate consistency scores for uniqueness and transitivity for each LLM in order to assess their temporal consistency and its impact on accuracy.
Consistency is later enforced on the predictions with an Integer Linear Programming (ILP) method, revealing that solving the inconsistencies does not improve the F1 score.

Overall, our contributions are:
\begin{itemize}
\item To the best of our knowledge, this is the first study of zero-shot BioTempRE. 
\item We provide extensive quantitative results of two types of prompts and five different LLMs.
\item We perform a novel temporal consistency analysis by calculating consistency scores of temporal properties.
\item We study how temporal consistency relates to accuracy and enforce it using an ILP method.
\item The code and data containing the prompts, the raw and the processed responses by the LLMs for around 600,000 pair instances, will be publicly shared for further analysis.\footnote{\url{https://github.com/vasilikikou/consistent_bioTempRE}}
\end{itemize}

\section{Related Work}

\subsection{Temporal Relation Extraction}

Multiple studies in TempRE have applied temporal properties to classifiers' predictions, either during training or at inference time, aiming to improve their performance \cite{Tang2013,Chambers2014,Ning2017,Ning2018,Wang2022}. Other works have also employed linguistic properties or properties based on causality \cite{Chambers2014,Ning2018}. \citet{Ning2018} formulated temporal, causal, and linguistic properties as constraints for an ILP method. Later, \citet{Liu2021} showed that ILP constraints can improve temporal consistency, although in certain cases, the F1 score may decrease.

\paragraph{Temporal Relation Extraction in the Medical Domain.}

The 2012 Informatics for Integrating Biology and the Bedside (i2b2) challenge was the first to address the BioTempRE task \cite{Sun2013}. 
The best-performing method involved merging predictions from different SVM and CRF classifiers with regard to temporal consistency \cite{Tang2013}.
Following challenges at SemEval, called Clinical TempEval, were organized from 2015 to 2017 \cite{Bethard2015,Bethard2016,Bethard2017} and utilized the THYME corpus \cite{Styler2014}.\footnote{The i2b2 dataset is publicly available. The THYME corpus is provided upon request, however our requests were not answered.} 
In 2015 and 2016, the best-performing methods were CRF- and SVM- based \cite{Velupillai2015,Lee2016,Khalifa2016}, while in 2017 the winning approach employed an LSTM \cite{Tourille2017}. Following approaches have utilized BERT \cite{Lin2019bert,Haq2021,Tu2023} for relation classification given a text and an event pair. Recently, \citet{Knez2024} introduced a multimodal method in which, they construct a graph with medical information and then, they combine textual representations (extracted by BERT) and graph representations (extracted by a GNN). Even though temporal consistency has been used in existing TempRE works, it has not been utilized for analyzing the performance of a model by calculating consistency scores.

\subsection{Zero-Shot Temporal Relation Extraction}

Zero-shot learning \citep{8413121} enables models to execute tasks without explicit training, a capability demonstrated by scaling models since GPT-3 \citep{NEURIPS2020_1457c0d6,wei2022finetuned}. 
Instruction tuning techniques \cite{wei2022finetuned} further enhance zero-shot learning in LLMs.
Recent openly available LLMs like LLama \cite{touvron2023llama} and Mixtral \cite{jiang2024mixtral} narrow the gap with closed-source models, while chain-of-thought (CoT) prompting \cite{NEURIPS2022_9d560961} has enhanced their ability to handle complex tasks. 
Research studies have shown that the temporal reasoning tasks remain challenging for LLMs \cite{Jain2023}, and specifically for TempRE, where \citet{Yuan2023} explored zero-shot TempRE with ChatGPT and found that it yields a large performance gap compared to supervised methods. However, previous research has not analyzed zero-shot TempRE in the medical domain or the temporal consistency and its impact on the performance of zero-shot TempRE - both gaps we aim to fill in our work. In this paper, we calculate consistency scores and study their connection to the F1 scores.

\section{Methodology}
\label{sec:method}

\subsection{Problem Formulation}

Given a text document $D$ and a set of events $E = \{e_1,..,e_{|E|}\}$ mentioned in the text, we create pairs of events, which are represented by the set $P = \{p_1,..,p_i,..,p_{|P|}\}$, where $p_i$ indicates the $i^{th}$ pair, $1\leq i \leq |P|$.
BioTempRE aims at assigning the appropriate temporal relation $r$ to the corresponding pair of events. Each $p_i \in P$ is described by two distinct events $\{e_j, e_k\}$, where $1\leq j, k \leq |E|$. Furthermore, each event $e \in E$ is characterized by the points in time at which it began and finished. 
These temporal points are denoted as $b$ and $f$, respectively. 

Following the work of \citet{Ning2018}, we employ the same relation scheme, which consists of $5$ different types of temporal relations $r$: \textit{before}, \textit{after}, \textit{includes}, \textit{is included}, and \textit{simultaneously}, represented by the label set $R_T = \{r_B,r_A,r_I,r_{II},r_S\}$. We choose this set of relations based on the fact that they are fine-grained and well-defined, and hence, suitable for creating temporal rules for our analysis.
An $r_B$ temporal relation indicates that $b(e_j) < b(e_k)$ and $f(e_j) < f(e_k)$ , while an $r_A$ temporal relation signifies that $b(e_j) > b(e_k)$ and $f(e_j) > f(e_k)$. Furthermore, $r_I$ indicates that $b(e_j) \leq b(e_k)$ and that $f(e_j) \geq f(e_k)$, and $r_{II}$ signifies that $b(e_j) \geq b(e_k)$ and that $f(e_j) \leq f(e_k)$. Finally, $r_S$ signifies that $b(e_j) = b(e_k)$ and $f(e_j) = f(e_k)$.

\subsection{Zero-shot BioTempRE}

We experimented with two different types of prompting: \textit{Batch-of-Questions} (BatchQA) and \textit{Chain-of-Thought} (CoT) \cite{NEURIPS2022_9d560961,Yuan2023} (see Figure~\ref{fig:prompts} in Appendix \ref{appendix}). In both, we start with a preamble consisting of the document text ($D$) and an instruction. Then, we introduce questions regarding the temporal relations for a pair of events $p_i$ consisting of events $e_j$ and $e_k$.\footnote{The questions were ordered randomly.} We formulated the question for each relation based on its temporal definition, as follows:
\begin{itemize}
    \item \textit{BEFORE}: Did $e_j$ start before $e_k$ started and end before $e_k$ ended?
    \item \textit{AFTER}: Did $e_j$ start after $e_k$ started and end after $e_k$ ended?
    \item \textit{INCLUDES}: Did $e_k$ start and end while $e_j$ was happening?
    \item \textit{IS INCLUDED}: Did $e_j$ start and end while $e_k$ was happening?
    \item \textit{SIMULTANEOUS}: Did $e_j$ and $e_k$ start and end at the same time?
\end{itemize}
We also specify the desired output format by adding \textit{``Answer with Yes or No''} to the end of each question.
For each event pair there is an independent interaction with the LLM, and depending on the type of prompt the questions mentioned above is sent to the LLM in one or multiple prompts.

\paragraph{Batch-of-Questions (BatchQA).} In BatchQA, a single prompt is sent to the LLM. In the preamble, after the document $D$, this instruction follows: \textit{``Given document $D$, answer the following questions ONLY with Yes or No.''}. Next, all the questions regarding the temporal relations are added in the same prompt. The expected model response includes five \textit{Yes/No} answers for each of the questions.

\paragraph{Chain-of-Thought (CoT).} The CoT approach is based on the chain-of-thought method proposed by \citet{Yuan2023} and follows a chat-like style consisting of multiple prompts. The first prompt is the preamble composed of the document $D$ and the question \textit{``Given the document $D$, are $e_j$ and $e_k$ referring to the same event? Answer ONLY with Yes or No.''}.
If the response is \textit{No}, then the questions are sent, each one in a separate prompt as they are mentioned above. 
If the response is \textit{Yes}, the phrase \textit{``In that event,''} is appended in the beginning of each question.

\section{Experimental Setup}

\subsection{Data}
\label{sec:data}

In our experiments, we use the dataset created for the 2012 i2b2 challenge, which consisted of 310 discharge summaries, 190 for training and 120 for testing. The texts were initially annotated with 8 fine-grained relations but due to low inter-annotator agreement these relations were merged to the following three: \textit{before, after and overlap}. Each discharge text contains 30.8 sentences on average, with each sentence having an average number of 17.7 tokens. The average number of tokens per discharge text is 514. 

The i2b2 dataset contains three types of events: 1) medical events, 2) time expressions, and 3) the dates of admission and discharge. The average number of medical events per discharge summary is 86.7, while the average number of time expressions is 10.5. 
The admission and discharge dates are included in each text; however, in a few cases, one of them might be missing. The annotators of i2b2 have assigned temporal relations to 27,540 pairs of events (gold pairs).

An important step in TempRE is to identify the pairs of events for which the models will decide if there is a relation expressed or not since it would not be feasible to check for every pair of events mentioned in a document. In order to generate candidate event pairs, we follow the approach of the best-performing method in the i2b2 challenge \cite{Tang2013}. This is a rule-based approach, which creates pairs consisting of every event and the admission and discharge dates, every two consecutive events within the same sentence, and events in the same as well as in different sentences based on linguistic rules. The generated candidate pairs are 60,840 in total, from which 28.16\% appears also in the gold pairs.

The five relations we use in our experiments (see Section \ref{sec:method}) are different from the gold ones existing in the dataset. In order to evaluate the prediction of our methods, we map the five relations to the three gold ones as follows: before $\rightarrow$ before, after $\rightarrow$ after, includes $\rightarrow$ overlap, is included $\rightarrow$ overlap and simultaneously $\rightarrow$ overlap.

\subsection{Methods}

\paragraph{LLMs}

We employed the following five (one closed-source and four open-weight) models of various sizes: GPT-3.5 (``ChatGPT''),\footnote{\url{https://openai.com/index/gpt-3-5-turbo-fine-tuning-and-api-updates/}} Gemma 7B \cite{gemmateam2024gemma},\footnote{\url{https://huggingface.co/google/gemma-1.1-7b-it}} Mixtral 8x7B \cite{jiang2024mixtral},\footnote{\url{https://huggingface.co/mistralai/Mixtral-8x7B-Instruct-v0.1}} Llama2 70B \cite{touvron2023llama},\footnote{\url{https://huggingface.co/meta-llama/Llama-2-70b-chat-hf}} and PMC-LLaMA 13B, which was pre-trained on medical text \cite{wu2023pmcllama}.\footnote{\url{https://huggingface.co/axiong/PMC\_LLaMA\_13B}} PMC-LLaMA was only instruction-tuned on QA data (respond to one question at a time) and struggled to follow the format of BatchQA prompts. Therefore, it was possible to use it only for CoT. The experiments were costly in terms of time (and money for GPT-3.5), especially for CoT where each question is sent separately. The running times range from three hours (Gemma BatchQA) to 7 days (Llama CoT) (see more details in Appendix \ref{appendix}).

\paragraph{Baselines}
We implemented a rule-based baseline, called W-order, where only the \textit{before} and \textit{after} relations are predicted for each event pair based on the order in which the events are mentioned in the text. A combination of the predictions of each LLM with the W-order predictions was also implemented. In cases where the LLM gave an negative or uncertain prediction for all the relations, the prediction of W-order was used instead.

\begin{table*}[t!]
\centering
    \begin{tabular}{l l c c c | c c c} \hline
    \toprule
    \multirow{2}{*}{\textbf{Setting}} & \multirow{2}{*}{\textbf{Method}} & \multicolumn{3}{c}{\textbf{Gold}} & \multicolumn{3}{c}{\textbf{Candidate}} \\ \cline{3-8}
    & & \textbf{P} & \textbf{R} & \textbf{F1} & \textbf{P} & \textbf{R} & \textbf{F1} \\ \hline
    Rule-based & W-order & 0.348 & 0.348 & 0.348 & 0.382 & 0.305 & 0.339 \\ \hline
    \multirow{3}{*}{Supervised} & \citet{Haq2021}\textsuperscript{\textdagger} & - & - & - & - & - & 0.736 \\
     & \citet{Tu2023}\textsuperscript{\textdagger} & - & - & - & 0.645 & 0.672 & 0.650 \\
     & \citet{Knez2024} & - & - & \textbf{0.820} & - & - & - \\
     & \citet{Tang2013} & - & - & - & 0.714 & 0.673 & \textbf{0.693} \\ \hline 
    \multirow{16}{*}{Zero-shot} & GPT-3.5 BatchQA & 0.588 & 0.083 & 0.132 & 0.607 & 0.060 & 0.101 \\ 
     & GPT-3.5 BatchQA + W-order & 0.395 & 0.397 & 0.396 & 0.424 & 0.340 & 0.377 \\
     & GPT-3.5 CoT & 0.400 & 0.641 & 0.491 & 0.387 & 0.494 & 0.432 \\  
     & GPT-3.5 CoT + W-order & 0.400 & 0.677 & 0.502 & 0.390 & 0.528 & 0.447 \\
     & Mixtral BatchQA & 0.458 & 0.534 & 0.491 & 0.420 & 0.392 & 0.404 \\ 
     & Mixtral BatchQA + W-order & 0.452 & 0.572 & 0.504 & 0.422 & 0.428 & 0.424 \\ 
     & Mixtral CoT & \textbf{0.681} & 0.504 & 0.576 & \textbf{0.694} & 0.422 & 0.520 \\
     & Mixtral CoT + W-order & 0.545 & 0.596 & 0.569 & 0.561 & 0.494 & 0.524 \\
     & Llama BatchQA & 0.366 & 0.371 & 0.367 & 0.316 & 0.254 & 0.281 \\
     & Llama BatchQA + W-order & 0.367 & 0.411 & 0.387 & 0.327 & 0.292 & 0.308 \\
     & Llama CoT & 0.549 & 0.710 & 0.615 & 0.551 & 0.567 & 0.555 \\
     & Llama CoT + W-order & 0.534 & 0.742 & \textbf{0.620} & 0.538 & 0.595 & \textbf{0.564} \\
     & Gemma BatchQA & 0.426 & 0.837 & 0.564 & 0.425 & 0.667 & 0.519 \\
     & Gemma BatchQA + W-order & 0.426 & \textbf{0.838} & 0.565 & 0.425 & \textbf{0.668} & 0.519 \\ 
     & Gemma CoT & 0.429 & 0.398 & 0.401 & 0.449 & 0.318 & 0.358 \\
     & Gemma CoT + W-order & 0.385 & 0.552 & 0.452 & 0.407 & 0.458 & 0.429 \\ 
     & PMC-LLaMA CoT & 0.395 & 0.516 & 0.439 & 0.406 & 0.425 & 0.408 \\ 
     & PMC-LLaMA CoT + W-order & 0.390 & 0.574 & 0.463 & 0.403 & 0.476 & 0.435 \\ \hline
\end{tabular}
\caption{Precision (P), recall (R) and F1 scores of TempRE methods on the gold and candidate pairs. Methods with \textsuperscript{\textdagger} use a different candidate pair generation than ours, their results are not directly comparable to ours.}
\label{tab:eval_results}
\end{table*}

\section{Zero-shot TempRE results}

To evaluate the correctness of the predicted relations, we calculate the precision, recall and F1 scores. 
For each pair of events, we check if the predicted relation matches the gold relation. Hence, we calculate the triple match between the predictions and the ground truth. In Table~\ref{tab:eval_results}, the results for the gold and candidate pairs are presented. In order to perform a fair comparison, considering that not every candidate pair of events has a gold annotation (and therefore it is unknown whether a prediction is correct or wrong), we only evaluate those generated candidate pairs that also are contained in the gold pairs. (If a gold pair does not exist in the generated candidate pairs, there is no prediction for it, and that would affect the recall score negatively.) In the Supervised setting we show scores reported by the corresponding papers. \citet{Knez2024} do not mention event detection or candidate pair generation, hence we assume they used the gold pairs. On the other hand, we show the results from \citet{Haq2021} and \citet{Tu2023} in the Candidate column since they operate on events they have detected in the text.

Our experiments demonstrate that the best performing methods are the same for the gold and the candidate pairs. As expected, the F1 score of the methods when the candidate pairs are used is lower, mostly due to the decrease in recall. The best performing method is \textit{Llama CoT + W-order} in terms of F1 score. On the other hand, \textit{Mixtral CoT} achieves the best precision score and \textit{Gemma BatchQA + W-order} the best recall. 
Overall, the supervised methods consistently outperform the methods in the zero-shot setting, with an avarage difference of approximately 20\% F1 score.
In general, most LLMs (except for \textit{Gemma}) exhibit improved performance when the CoT prompting approach is used.
However, in an LLM-based comparison, we observe that the performance varies depending on the type of prompt used. For example, Llama with CoT has the highest F1 score, but when BatchQA is used, the score drops almost in half.
Moreover, the combination of \textit{W-order} predictions with the zero-shot methods yields improvements in recall and F1 score, but in most cases, it harms precision. Notably, \textit{PMC-LLaMA}, the medical LLM we employed, has low results and is often outperformed by the general domain LLMs, showing no advantage from pre-training on biomedical text.

\section{Temporal consistency analysis}

Considering the temporal nature of the TempRE task, we investigate the impact of incorporating the following two temporal properties in the zero-shot setting: 1. \textit{uniqueness}, requiring that each event pair has exactly one relation, and 2. \textit{transitivity}. First, we evaluate the zero-shot methods based on their consistency, i.e., if their predictions follow the temporal properties or not. Then, we use ILP to enforce temporal consistency on the predictions. Specifically we examine the following three questions: 
\begin{itemize}
    \item How consistent are the different zero-shot methods?
    \item How is the temporal consistency of the predictions connected to their correctness?
    \item Can the predictions be improved by a temporal constraint-based ILP method?
\end{itemize}

\begin{table*}[h!]
\centering
  \begin{tabular}{l c c c c | c c c c} \hline
    \multirow{2}{*}{\bf Method} & \multicolumn{4}{c}{\textbf{Gold}} & \multicolumn{4}{c}{\textbf{Candidate}} \\ \cline{2-9} 
    & \bf $c_U$ (\%) & \bf \# 0 & \bf \# >1 & \bf $c_T$ (\%) & \bf $c_U$ (\%) & \bf \# 0 & \bf \# >1 & \bf $c_T$ (\%) \\ \hline
    GPT-3.5 BatchQA & 8.03 & 24,476 & 860 & 70.34 & 5.06 & 56,457 & 1,306 & 68.78 \\
    GPT-3.5 CoT & 13.07 & 2,657 & 21,284 & 46.58 & 14.91 & 6,194 & 45,573 &  47.29 \\
    Mixtral BatchQA & 59.94 & 3,102 & 7,931 & 71.20 & 60.13 & 7,192 & 17,063 & 71.87 \\
    Mixtral CoT & 37.60 & 10,315 & 6,868 & 68.99 & 35.22 & 27,434 & 11,980 &  67.44 \\
    Llama BatchQA & \textbf{71.67} & 2,858 & 4,945 & \textbf{82.35}  & \textbf{70.58} & 6,451 & 11,446 & \textbf{80.05} \\
    Llama CoT & 30.55 & 2,916 & 16,211 & 60.39 & 33.64 & 6,864 & 33,507 &  59.45 \\
    Gemma BatchQA & 2.67 & 57 & 26,747 & 63.04 & 2.26 & 115 & 59,347 & 62.59 \\
    Gemma CoT & 3.82 & 14,159 & 12,335 & 60.00 & 3.00 & 32,605 & 26,411 & 60.56 \\
    PMC-LLaMA CoT & 33.18 & 7,469 & 10,933 & 60.45 & 31.88 & 17,977 & 23,465 & 59.85 \\ \hline
    W/ ILP reasoning & 100 & 0 & 0 & 100 & 100 & 0 & 0 & 100 \\ \hline
\end{tabular}
\caption{Temporal consistency scores for uniqueness ($c_U$) and transitivity ($c_T$) for each model. The consistency scores show the percentage of pairs which are consistent for the corresponding property. \# 0 and \# >1 shows the number of pairs with none and more than one predictions respectively.}
\label{tab:consistency}
\end{table*}

\paragraph{How consistent are the different zero-shot methods?}
We calculate two consistency scores: one for uniqueness $c_U$ and one for transitivity $c_T$, which show the percentage of cases where the corresponding temporal property was not violated. The consistency score for uniqueness is calculated based on the number of pairs as follows:
\begin{equation}
    c_U = \frac{\sum_{i=1}^P p_{i, |r_i| = 1}}{|P|},
\end{equation}
where only the pairs $p_i$ with a singular predicted relation $r_i$ are considered.
In Table~\ref{tab:consistency}, the consistency scores for uniqueness are reported. Furthermore, we present the number of pairs for which no relation was predicted (\# 0) and the number of pairs with more than one predicted relation (\# >1).
We observe that all the models struggle to keep consistency, especially because of assigning more than one relation to a pair. For the majority of the evaluated LLMs, this occurs for at least 50\% of the pairs and can go up to 97\% (Gemma BatchQA). In this evaluation, we also find that there is no clear winner among the LLMs or the prompt types, since the same LLM shows different levels of consistency with a different prompt type. The combination with the highest consistency for uniqueness is Llama with BatchQA.

The consistency score for transitivity is calculated based on triples of event pairs in the following form:
$((e_i, e_j), (e_j, e_k), (e_i, e_k))$. We first find these triples in the dataset and then obtain the relations predicted for them. If $r_1$, $r_2$ and $r_3$ are the predictions for each respective pair in the triple, then for $r_3$, it should hold that $r_3 \in trans(r_1, r_2)$.\footnote{The transitive relations for the relation set we used can be found in Table~\ref{tab:trans_relations} in Appendix \ref{appendix}.} If it does not hold, then we have a transitivity violation. Therefore $c_T$ is calculated as:
\begin{equation}
    c_T = \frac{\sum_{i=1}^{|Tr|} t_{i, r_3 \in trans(r_1, r_2)}}{|Tr|},
\end{equation}
where $Tr$ is the set of transitivity triples and, for each triple $t_i$, the transitivity for the predicted relations holds.\footnote{Triples where at least one pair was not assigned a relation are excluded from this calculation.}Table~\ref{tab:consistency} showcases the $c_T$ scores for each of the evaluated methods. Similar to \textit{uniqueness}, Llama BatchQA demonstrates the highest consistency for \textit{transitivity}. In general, for all LLMs, we observe that the BatchQA approach yields higher transitivity consistency scores than CoT.

\begin{table*}[t!]
\centering
    \begin{tabular}{l c c c c c c} \hline
    \toprule
    \multirow{2}{*}{\textbf{Method}} & \multicolumn{3}{c}{\textbf{W/o ILP reasoning}} & \multicolumn{3}{c}{\textbf{W/ ILP reasoning}} \\ \cline{2-7} 
    & \textbf{P} & \textbf{R} & \textbf{F1} & \textbf{P} & \textbf{R} & \textbf{F1} \\ \hline
    Llama BatchQA & 0.366 & 0.371 & 0.367 & 0.366 & 0.366 & 0.366 \\
    Llama CoT & 0.549 & 0.710 & 0.615 & 0.549 & 0.549 &  0.549 \\\hline
\end{tabular}
\caption{Precision (P), recall (R) and F1 scores before and after the ILP temporal reasoning step on the gold pairs.}
\label{tab:ILP_results}
\end{table*}

\paragraph{How is the temporal consistency of the predictions connected to their correctness?}

When comparing the consistency scores with F1, we observe a contradiction. Models that have high consistency have a lower F1 score. In particular, Llama BatchQA is the most consistent in terms of uniqueness and transitivity, but has one of the lowest F1 scores. Especially for the candidate pairs, the F1 score is even lower than the rule-based baseline, yet $c_U$ is 70.58\% and $c_T$ is 80.05\%. Moreover, Llama CoT, which is the best in terms of F1 score, has low consistency with around only 30\% of predictions being unique and 60\% correct transitivity triples.
These insights suggest that temporal consistency does not always mean correctness.

\paragraph{Can the predictions be improved by a temporal constraint-based ILP approach?}
Following the approach of \citet{Ning2017,Ning2018}, we implemented an ILP step that uses the temporal properties as constraints and changes inconsistent predictions so that the constraints are not violated.\footnote{For the ILP implementation we used the Gurobi optimizer (\url{https://www.gurobi.com/solutions/gurobi-optimizer/}).} With this study we examine whether consistency can improve the predictions. First, we assign a confidence score $sc$ to each triple $(e_i, e_j, r_k), \forall r_k \in R_T$. The score $sc$ for a pair of events $p = (e_i, e_j)$ equals 1, if the relation was predicted from the model, and 0.2 otherwise. 
Next, we create a binary vector, which is optimized with ILP. We refer to it as the indicator $I(p_i, r_i) \in [0,1], \forall p \in P, r \in R_T$.
We formulate the constraints as follows:
\begin{itemize}
    \item Uniqueness: 
    \begin{equation}
   \sum_{p \in P, r \in R_T} I(p, r) = 1 
    \end{equation}
    \item Symmetry:
    \begin{equation}
    I(p_i, r_i) = I(p^s_i, \bar{r}_i)
    \end{equation}
    where $p_i = (e_i, e_j)$ and $p^s_i = (e_j, e_i)$, and $\bar{r}_i$ is the reverse relation of $r_i$.\footnote{The reverse of each relation can be found in Table \ref{tab:sym_relations} in Appendix \ref{appendix}.}
    \item Transitivity:
    \begin{equation}
    I((e_i, e_j), r_1) + I((e_j, e_k), r_2) - \sum_{r_3 \in tr(r_1,r_2)} \leq 1
    \end{equation}

    where $r_1, r_2, r_3 \in R_T$ and $trans(r_1, r_2)$ is the set of relations that are the transitive of relations $r_1$ and $r_2$.
\end{itemize}

The objective of the ILP method is to maximize the confidence score based on the indicator:
\begin{equation}
    \overset{\wedge}{I} = \argmax \sum_{p \in P} \sum_{r \in R_T} sc(p, r) I(p, r)
\end{equation}

\begin{figure*}
\includegraphics[width=\textwidth]{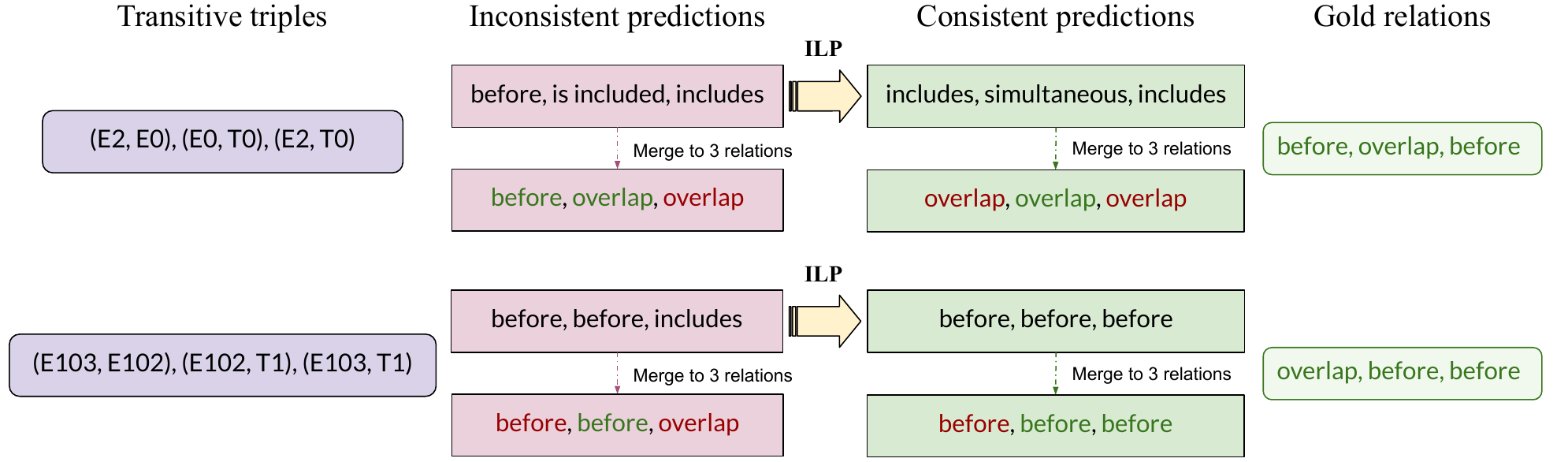}
    \caption{Examples of two transitive triples with inconsistent predictions. After the ILP the predictions are consistent but still different from the gold relations.}
    \vspace{-12pt}
    \label{fig:examples}
\end{figure*}

As shown in Table~\ref{tab:consistency}, when the ILP reasoning step is applied, the consistency scores for both uniqueness and transitivity reach 100\%.
We applied this step to the predictions of Llama BatchQA and Llama CoT, which are the models with the highest contradiction between F1 and consistency.
In Table~\ref{tab:ILP_results}, we show the results before and after applying the temporal constraints. Even though the consistency of the predictions after reasoning is 100\%, the F1 score decreases slightly for BatchQA and by 0.066 for CoT. This means that the predictions are temporally consistent, but they are not accurate. To get a better understanding of this issue, Figure~\ref{fig:examples} demonstrates two examples of transitivity triples for which the predictions violate the transitivity constraint. This indicates that at least one of the three predictions is incorrect and needs to change. In the top example, the first two relations were correct, but these relations were changed by the ILP step, resulting in only one relation being correct in the consistent predictions. In the bottom example, only one relation was changed to enforce consistency. This resulted in two correctly predicted relations after the ILP, but still the first relation remained incorrect. This analysis highlights our previous observation regarding the relation between consistency and accuracy, and points out to the need of a better alignment of these two aspects in order for models to achieve a better temporal reasoning.

\section{Pairs distance-based analysis}

\begin{figure}
\includegraphics[scale=0.23]{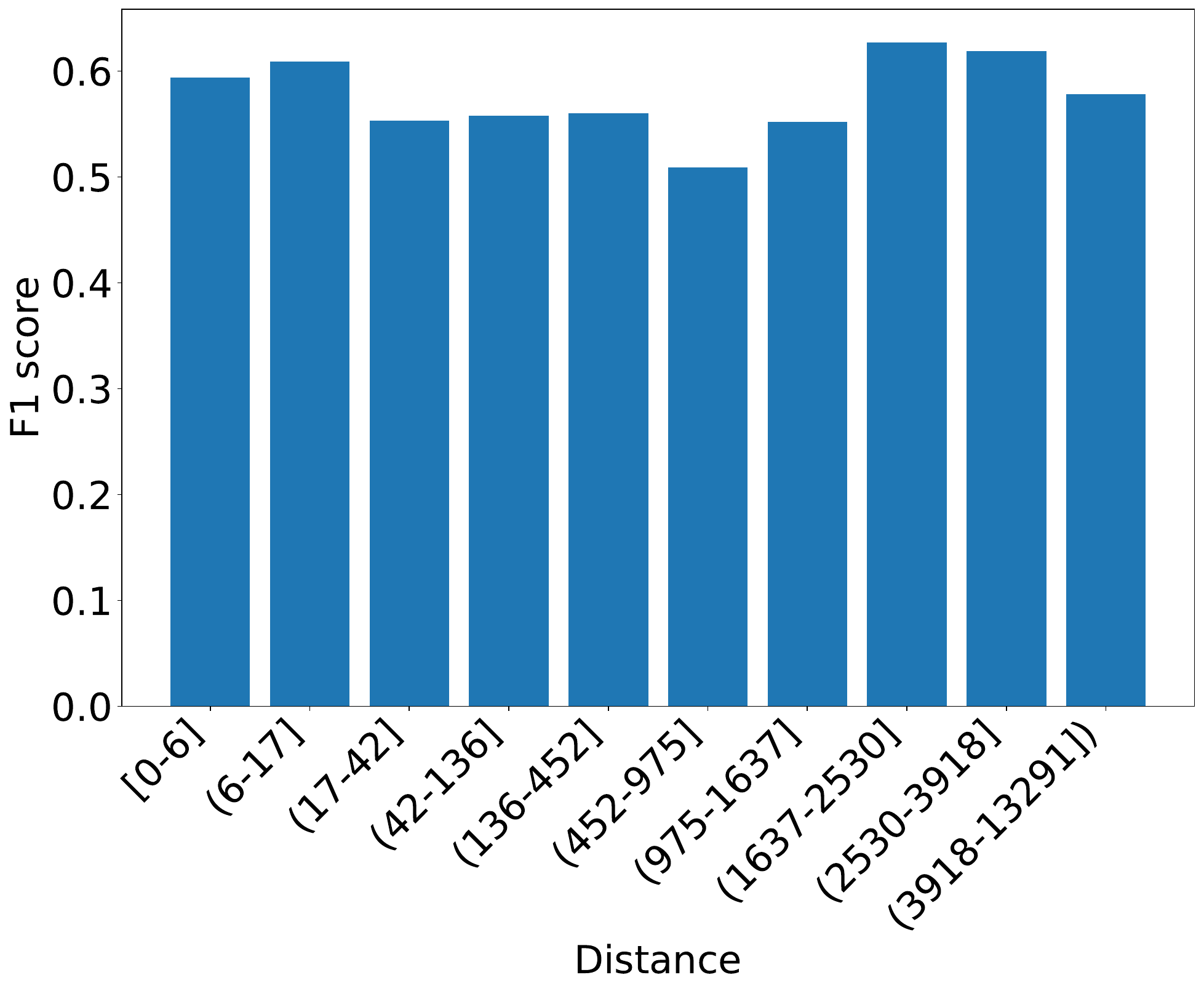}
    \caption{Barplot where each bar represents a range of distances between events in the gold pairs. The y axis shows the F1 score of the predictions for the pairs in each bar.}
    \vspace{-12pt}
    \label{fig:distances}
\end{figure}

Since clinical notes contain long texts (see Section~\ref{sec:data}), we perform an analysis based on the distance of event pairs for the best-performing LLM (Llama CoT). First, we calculate the distance in terms of characters between the events for all the gold pairs. Then, we sort the pairs by their distances and split them to 10 bins, so that each bin contains roughly the same amount of pairs. Finally, the F1 score is calculated for the prediction of the pairs contained in each bin. Figure~\ref{fig:distances} depicts the barplot with the bars representing the pairs in the specific distance range and the their corresponding F1 scores. We observe that 37.5\% of the pairs have a distance of 0 to 100 characters. Larger distances appear less frequently and hence the range of distance is smaller in the first bars, while the last bars have larger ranges. 
There is no consistent decrease in F1 score as the distance increases, meaning that the model is not affected by the distance of events in the text.

\section{Conclusion}

BioTempRE is an important task that tackles medical event ordering, with beneficial applications in healthcare. In this paper, we performed BioTempRE on clinical notes in a zero-shot setting employing five different LLMs. Two types of prompts were used, namely BatchQA and CoT, in order to obtain LLMs' responses. The zero-shot performance of all LLMs is lower compared to supervised learning methods. Moreover, we perform a temporal evaluation by calculating the consistency score of each LLM for the temporal properties of uniqueness and transivity. We find that, in general, LLMs' predictions are temporally inconsistent and, interestingly, the model with the higher consistency scores (Llama BatchQA) has one of the lowest F1 scores. An ILP method utilized to enforce consistency on the models' predictions fails to improve their accuracy. These findings indicate the importance of the relation between temporal consistency and correctness, emphasizing the need for further study in order to assist in temporal reasoning.

\section*{Acknowledgments}
This research has been funded by the Vienna Science and Technology Fund (WWTF)[10.47379/VRG19008] ”Knowledge-infused Deep Learning for Natural Language Processing”.

\bibliography{custom}

\begin{thebibliography}{36}
\providecommand{\natexlab}[1]{#1}

\bibitem[{Allen(1983)}]{Allen1983}
James~F Allen. 1983.
\newblock Maintaining knowledge about temporal intervals.
\newblock \emph{Communications of the ACM}, 26(11):832--843.

\bibitem[{Bethard et~al.(2015)Bethard, Derczynski, Savova, Pustejovsky, and Verhagen}]{Bethard2015}
Steven Bethard, Leon Derczynski, Guergana Savova, James Pustejovsky, and Marc Verhagen. 2015.
\newblock Semeval-2015 task 6: Clinical tempeval.
\newblock In \emph{proceedings of the 9th International Workshop on Semantic Evaluation (SemEval 2015)}, pages 806--814.

\bibitem[{Bethard et~al.(2016)Bethard, Savova, Chen, Derczynski, Pustejovsky, and Verhagen}]{Bethard2016}
Steven Bethard, Guergana Savova, Wei-Te Chen, Leon Derczynski, James Pustejovsky, and Marc Verhagen. 2016.
\newblock Semeval-2016 task 12: Clinical tempeval.
\newblock In \emph{Proceedings of the 10th international workshop on semantic evaluation (SemEval-2016)}, pages 1052--1062.

\bibitem[{Bethard et~al.(2017)Bethard, Savova, Palmer, and Pustejovsky}]{Bethard2017}
Steven Bethard, Guergana Savova, Martha Palmer, and James Pustejovsky. 2017.
\newblock \href {https://doi.org/10.18653/v1/S17-2093} {{S}em{E}val-2017 task 12: Clinical {T}emp{E}val}.
\newblock In \emph{Proceedings of the 11th International Workshop on Semantic Evaluation ({S}em{E}val-2017)}, pages 565--572, Vancouver, Canada.

\bibitem[{Brown et~al.(2020)Brown, Mann, Ryder, Subbiah, Kaplan, Dhariwal, Neelakantan, Shyam, Sastry, Askell, Agarwal, Herbert-Voss, Krueger, Henighan, Child, Ramesh, Ziegler, Wu, Winter, Hesse, Chen, Sigler, Litwin, Gray, Chess, Clark, Berner, McCandlish, Radford, Sutskever, and Amodei}]{NEURIPS2020_1457c0d6}
Tom Brown, Benjamin Mann, Nick Ryder, Melanie Subbiah, Jared~D Kaplan, Prafulla Dhariwal, Arvind Neelakantan, Pranav Shyam, Girish Sastry, Amanda Askell, Sandhini Agarwal, Ariel Herbert-Voss, Gretchen Krueger, Tom Henighan, Rewon Child, Aditya Ramesh, Daniel Ziegler, Jeffrey Wu, Clemens Winter, Chris Hesse, Mark Chen, Eric Sigler, Mateusz Litwin, Scott Gray, Benjamin Chess, Jack Clark, Christopher Berner, Sam McCandlish, Alec Radford, Ilya Sutskever, and Dario Amodei. 2020.
\newblock \href {https://proceedings.neurips.cc/paper_files/paper/2020/file/1457c0d6bfcb4967418bfb8ac142f64a-Paper.pdf} {Language models are few-shot learners}.
\newblock In \emph{Advances in Neural Information Processing Systems}, volume~33, pages 1877--1901. Curran Associates, Inc.

\bibitem[{Bubeck et~al.(2023)Bubeck, Chandrasekaran, Eldan, Gehrke, Horvitz, Kamar, Lee, Lee, Li, Lundberg et~al.}]{Bubeck2023}
S{\'e}bastien Bubeck, Varun Chandrasekaran, Ronen Eldan, Johannes Gehrke, Eric Horvitz, Ece Kamar, Peter Lee, Yin~Tat Lee, Yuanzhi Li, Scott Lundberg, et~al. 2023.
\newblock Sparks of artificial general intelligence: Early experiments with gpt-4.
\newblock \emph{arXiv preprint arXiv:2303.12712}.

\bibitem[{Chambers et~al.(2014)Chambers, Cassidy, McDowell, and Bethard}]{Chambers2014}
Nathanael Chambers, Taylor Cassidy, Bill McDowell, and Steven Bethard. 2014.
\newblock Dense event ordering with a multi-pass architecture.
\newblock \emph{Transactions of the Association for Computational Linguistics}, 2:273--284.

\bibitem[{Gumiel et~al.(2021)Gumiel, Silva~e Oliveira, Claveau, Grabar, Paraiso, Moro, and Carvalho}]{Gumiel2021survey}
Yohan~Bonescki Gumiel, Lucas~Emanuel Silva~e Oliveira, Vincent Claveau, Natalia Grabar, Emerson~Cabrera Paraiso, Claudia Moro, and Deborah~Ribeiro Carvalho. 2021.
\newblock Temporal relation extraction in clinical texts: a systematic review.
\newblock \emph{ACM Computing Surveys (CSUR)}, 54(7):1--36.

\bibitem[{Haq et~al.(2021)Haq, Kocaman, and Talby}]{Haq2021}
Hasham~Ul Haq, Veysel Kocaman, and David Talby. 2021.
\newblock Deeper clinical document understanding using relation extraction.
\newblock \emph{arXiv preprint arXiv:2112.13259}.

\bibitem[{Jain et~al.(2023)Jain, Sojitra, Acharya, Saha, Jatowt, and Dandapat}]{Jain2023}
Raghav Jain, Daivik Sojitra, Arkadeep Acharya, Sriparna Saha, Adam Jatowt, and Sandipan Dandapat. 2023.
\newblock Do language models have a common sense regarding time? revisiting temporal commonsense reasoning in the era of large language models.
\newblock In \emph{Proceedings of the 2023 Conference on Empirical Methods in Natural Language Processing}, pages 6750--6774, Singapore.

\bibitem[{Jiang et~al.(2024)Jiang, Sablayrolles, Roux, Mensch, Savary, Bamford, Chaplot, Casas, Hanna, Bressand et~al.}]{jiang2024mixtral}
Albert~Q Jiang, Alexandre Sablayrolles, Antoine Roux, Arthur Mensch, Blanche Savary, Chris Bamford, Devendra~Singh Chaplot, Diego de~las Casas, Emma~Bou Hanna, Florian Bressand, et~al. 2024.
\newblock Mixtral of experts.
\newblock \emph{arXiv preprint arXiv:2401.04088}.

\bibitem[{Khalifa et~al.(2016)Khalifa, Velupillai, and Meystre}]{Khalifa2016}
Abdulrahman Khalifa, Sumithra Velupillai, and Stephane Meystre. 2016.
\newblock Utahbmi at semeval-2016 task 12: extracting temporal information from clinical text.
\newblock In \emph{Proceedings of the 10th International Workshop on Semantic Evaluation (SemEval-2016)}, pages 1256--1262.

\bibitem[{Knez and {\v{Z}}itnik(2024)}]{Knez2024}
Timotej Knez and Slavko {\v{Z}}itnik. 2024.
\newblock Multimodal learning for temporal relation extraction in clinical texts.
\newblock \emph{Journal of the American Medical Informatics Association}, page ocae059.

\bibitem[{Lee et~al.(2016)Lee, Xu, Wang, Zhang, Moon, Xu, and Wu}]{Lee2016}
Hee-Jin Lee, Hua Xu, Jingqi Wang, Yaoyun Zhang, Sungrim Moon, Jun Xu, and Yonghui Wu. 2016.
\newblock Uthealth at semeval-2016 task 12: an end-to-end system for temporal information extraction from clinical notes.
\newblock In \emph{Proceedings of the 10th International Workshop on Semantic Evaluation (SemEval-2016)}, pages 1292--1297.

\bibitem[{Lin et~al.(2019)Lin, Miller, Dligach, Bethard, and Savova}]{Lin2019bert}
Chen Lin, Timothy Miller, Dmitriy Dligach, Steven Bethard, and Guergana Savova. 2019.
\newblock A bert-based universal model for both within-and cross-sentence clinical temporal relation extraction.
\newblock In \emph{Proceedings of the 2nd Clinical Natural Language Processing Workshop}, pages 65--71.

\bibitem[{Liu et~al.(2021)Liu, Xu, Chen, and Zhang}]{Liu2021}
Jian Liu, Jinan Xu, Yufeng Chen, and Yujie Zhang. 2021.
\newblock Discourse-level event temporal ordering with uncertainty-guided graph completion.
\newblock In \emph{IJCAI}, pages 3871--3877.

\bibitem[{Ning et~al.(2017)Ning, Feng, and Roth}]{Ning2017}
Qiang Ning, Zhili Feng, and Dan Roth. 2017.
\newblock \href {https://doi.org/10.18653/v1/D17-1108} {A structured learning approach to temporal relation extraction}.
\newblock In \emph{Proceedings of the 2017 Conference on Empirical Methods in Natural Language Processing}, pages 1027--1037, Copenhagen, Denmark. Association for Computational Linguistics.

\bibitem[{Ning et~al.(2018)Ning, Feng, Wu, and Roth}]{Ning2018}
Qiang Ning, Zhili Feng, Hao Wu, and Dan Roth. 2018.
\newblock \href {https://doi.org/10.18653/v1/P18-1212} {Joint reasoning for temporal and causal relations}.
\newblock In \emph{Proceedings of the 56th Annual Meeting of the Association for Computational Linguistics (Volume 1: Long Papers)}, pages 2278--2288, Melbourne, Australia.

\bibitem[{Styler~IV et~al.(2014)Styler~IV, Bethard, Finan, Palmer, Pradhan, De~Groen, Erickson, Miller, Lin, Savova et~al.}]{Styler2014}
William~F Styler~IV, Steven Bethard, Sean Finan, Martha Palmer, Sameer Pradhan, Piet~C De~Groen, Brad Erickson, Timothy Miller, Chen Lin, Guergana Savova, et~al. 2014.
\newblock Temporal annotation in the clinical domain.
\newblock \emph{Transactions of the association for computational linguistics}, 2:143--154.

\bibitem[{Sun et~al.(2013)Sun, Rumshisky, and Uzuner}]{Sun2013}
Weiyi Sun, Anna Rumshisky, and Ozlem Uzuner. 2013.
\newblock Evaluating temporal relations in clinical text: 2012 i2b2 challenge.
\newblock \emph{Journal of the American Medical Informatics Association}, 20(5):806--813.

\bibitem[{Tan et~al.(2023)Tan, Ng, and Bing}]{Tan2023}
Qingyu Tan, Hwee~Tou Ng, and Lidong Bing. 2023.
\newblock Towards benchmarking and improving the temporal reasoning capability of large language models.
\newblock In \emph{Proceedings of the 61st Annual Meeting of the Association for Computational Linguistics Volume 1: Long Papers}, pages 14820--14835, Toronto, Canada.

\bibitem[{Tang et~al.(2013)Tang, Wu, Jiang, Chen, Denny, and Xu}]{Tang2013}
Buzhou Tang, Yonghui Wu, Min Jiang, Yukun Chen, Joshua~C Denny, and Hua Xu. 2013.
\newblock A hybrid system for temporal information extraction from clinical text.
\newblock \emph{Journal of the American Medical Informatics Association}, 20(5):828--835.

\bibitem[{Team et~al.(2024)Team, Mesnard, Hardin, Dadashi, Bhupatiraju, Pathak, Sifre, Rivi{\`e}re, Kale, Love et~al.}]{gemmateam2024gemma}
Gemma Team, Thomas Mesnard, Cassidy Hardin, Robert Dadashi, Surya Bhupatiraju, Shreya Pathak, Laurent Sifre, Morgane Rivi{\`e}re, Mihir~Sanjay Kale, Juliette Love, et~al. 2024.
\newblock Gemma: Open models based on gemini research and technology.
\newblock \emph{arXiv preprint arXiv:2403.08295}.

\bibitem[{Tourille et~al.(2017)Tourille, Ferret, N{\'e}v{\'e}ol, and Tannier}]{Tourille2017}
Julien Tourille, Olivier Ferret, Aur{\'e}lie N{\'e}v{\'e}ol, and Xavier Tannier. 2017.
\newblock \href {https://doi.org/10.18653/v1/P17-2035} {Neural architecture for temporal relation extraction: A {B}i-{LSTM} approach for detecting narrative containers}.
\newblock In \emph{Proceedings of the 55th Annual Meeting of the Association for Computational Linguistics (Volume 2: Short Papers)}, pages 224--230, Vancouver, Canada.

\bibitem[{Touvron et~al.(2023)Touvron, Martin, Stone, Albert, Almahairi, Babaei, Bashlykov, Batra, Bhargava, Bhosale et~al.}]{touvron2023llama}
Hugo Touvron, Louis Martin, Kevin Stone, Peter Albert, Amjad Almahairi, Yasmine Babaei, Nikolay Bashlykov, Soumya Batra, Prajjwal Bhargava, Shruti Bhosale, et~al. 2023.
\newblock Llama 2: Open foundation and fine-tuned chat models.
\newblock \emph{arXiv preprint arXiv:2307.09288}.

\bibitem[{Tu et~al.(2023)Tu, Han, and Nenadic}]{Tu2023}
Hangyu Tu, Lifeng Han, and Goran Nenadic. 2023.
\newblock Extraction of medication and temporal relation from clinical text using neural language models.
\newblock In \emph{2023 IEEE International Conference on Big Data (BigData)}, pages 2735--2744.

\bibitem[{Velupillai et~al.(2015)Velupillai, Mowery, Abdelrahman, Christensen, and Chapman}]{Velupillai2015}
Sumithra Velupillai, Danielle~L Mowery, Samir Abdelrahman, Lee Christensen, and Wendy Chapman. 2015.
\newblock Blulab: Temporal information extraction for the 2015 clinical tempeval challenge.
\newblock In \emph{Proceedings of the 9th International Workshop on Semantic Evaluation (SemEval 2015)}, pages 815--819.

\bibitem[{Wang et~al.(2022)Wang, Li, and Xu}]{Wang2022}
Liang Wang, Peifeng Li, and Sheng Xu. 2022.
\newblock Dct-centered temporal relation extraction.
\newblock In \emph{Proceedings of the 29th International Conference on Computational Linguistics}, pages 2087--2097.

\bibitem[{Wei et~al.(2022{\natexlab{a}})Wei, Bosma, Zhao, Guu, Yu, Lester, Du, Dai, and Le}]{wei2022finetuned}
Jason Wei, Maarten Bosma, Vincent Zhao, Kelvin Guu, Adams~Wei Yu, Brian Lester, Nan Du, Andrew~M. Dai, and Quoc~V Le. 2022{\natexlab{a}}.
\newblock \href {https://openreview.net/forum?id=gEZrGCozdqR} {Finetuned language models are zero-shot learners}.
\newblock In \emph{International Conference on Learning Representations}.

\bibitem[{Wei et~al.(2022{\natexlab{b}})Wei, Wang, Schuurmans, Bosma, ichter, Xia, Chi, Le, and Zhou}]{NEURIPS2022_9d560961}
Jason Wei, Xuezhi Wang, Dale Schuurmans, Maarten Bosma, brian ichter, Fei Xia, Ed~Chi, Quoc~V Le, and Denny Zhou. 2022{\natexlab{b}}.
\newblock \href {https://proceedings.neurips.cc/paper_files/paper/2022/file/9d5609613524ecf4f15af0f7b31abca4-Paper-Conference.pdf} {Chain-of-thought prompting elicits reasoning in large language models}.
\newblock In \emph{Advances in Neural Information Processing Systems}, volume~35, pages 24824--24837. Curran Associates, Inc.

\bibitem[{Wenzel and Jatowt(2023)}]{Wenzel2023}
Georg Wenzel and Adam Jatowt. 2023.
\newblock An overview of temporal commonsense reasoning and acquisition.
\newblock \emph{arXiv preprint arXiv:2308.00002}.

\bibitem[{Wu et~al.(2023{\natexlab{a}})Wu, Lin, Zhang, Zhang, Wang, and Xie}]{wu2023pmcllama}
Chaoyi Wu, Weixiong Lin, Xiaoman Zhang, Ya~Zhang, Yanfeng Wang, and Weidi Xie. 2023{\natexlab{a}}.
\newblock \href {https://arxiv.org/abs/2304.14454} {Pmc-llama: Towards building open-source language models for medicine}.
\newblock \emph{Preprint}, arXiv:2304.14454.

\bibitem[{Wu et~al.(2023{\natexlab{b}})Wu, Zhang, Cao, Yu, Dai, Ma, Liu, Zhao, Li, Liu et~al.}]{Wu2023}
Zihao Wu, Lu~Zhang, Chao Cao, Xiaowei Yu, Haixing Dai, Chong Ma, Zhengliang Liu, Lin Zhao, Gang Li, Wei Liu, et~al. 2023{\natexlab{b}}.
\newblock Exploring the trade-offs: Unified large language models vs local fine-tuned models for highly-specific radiology nli task.
\newblock \emph{arXiv preprint arXiv:2304.09138}.

\bibitem[{Xian et~al.(2019)Xian, Lampert, Schiele, and Akata}]{8413121}
Y.~Xian, C.~H. Lampert, B.~Schiele, and Z.~Akata. 2019.
\newblock \href {https://doi.org/10.1109/TPAMI.2018.2857768} {Zero-shot learning—a comprehensive evaluation of the good, the bad and the ugly}.
\newblock \emph{IEEE Transactions on Pattern Analysis \& Machine Intelligence}, 41(09):2251--2265.

\bibitem[{Yao et~al.(2022)Yao, Breitfeller, Naik, Zhou, and Rose}]{Yao2022}
Hao-Ren Yao, Luke Breitfeller, Aakanksha Naik, Chunxiao Zhou, and Carolyn Rose. 2022.
\newblock Multi-scale contrastive knowledge co-distillation for event temporal relation extraction.
\newblock \emph{arXiv preprint arXiv:2209.00568}.

\bibitem[{Yuan et~al.(2023)Yuan, Xie, and Ananiadou}]{Yuan2023}
Chenhan Yuan, Qianqian Xie, and Sophia Ananiadou. 2023.
\newblock \href {https://doi.org/10.18653/v1/2023.bionlp-1.7} {Zero-shot temporal relation extraction with {C}hat{GPT}}.
\newblock In \emph{The 22nd Workshop on Biomedical Natural Language Processing and BioNLP Shared Tasks}, pages 92--102, Toronto, Canada. Association for Computational Linguistics.

\end{thebibliography}

\appendix

\newpage
\section{Appendix}
\label{appendix}

\begin{figure*}
\includegraphics[width=\textwidth]{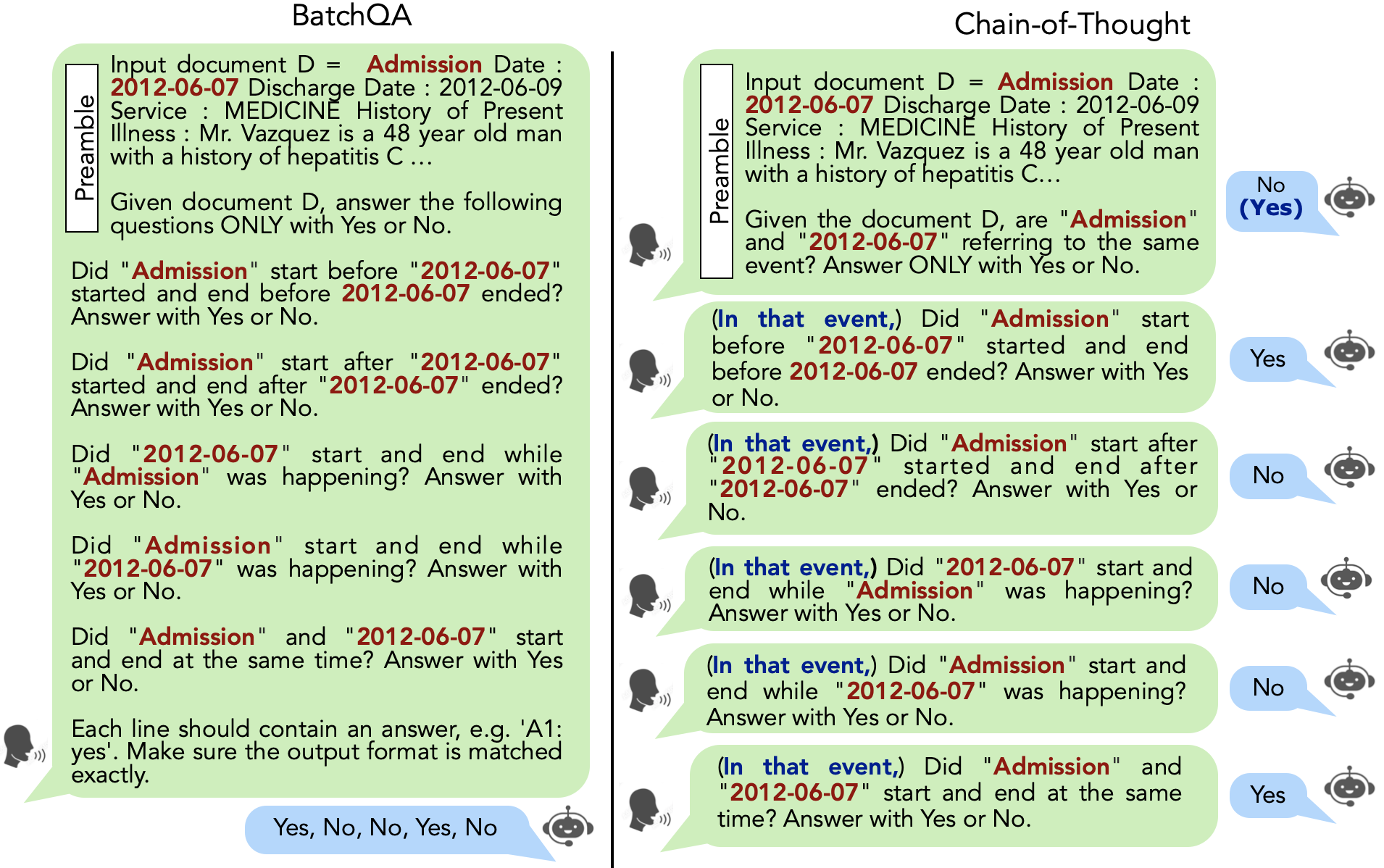}
    \vspace{-20pt}
    \caption{Examples of an interaction with the LLM using two different prompting strategies: BatchQA and Chain-of-Thought.}
    \vspace{-12pt}
    \label{fig:prompts}
\end{figure*}

\paragraph{Technical details}

Getting responses from GPT-3.5 for all the pairs for both types of prompts costed around 800\$ and lasted 27 hours.
For the open-source models we used a single H100 GPU, and for the rest two H100 GPUs.
The running time for each model was:
\begin{itemize}
    \item Mixtral 8x7B BatchQA: 6 hours
    \item Mixtral 8x7B CoT: 48 hours
    \item Llama2 70B BatchQA: 24 hours
    \item Llama2 70B CoT: 7 days
    \item Gemma 7B BatchQA: 3 hours
    \item Gemma 7B CoT: 25 hours
    \item PMC-Llama 13B CoT: 2.5 days
\end{itemize}

\newpage
\begin{table*}[h!]
\centering
  \begin{tabular}{c c c} \hline
    \bf $r_1$ & \bf $r_2$ & \bf $trans(r_1, r_2)$ \\ \hline
    before & before & before \\
    after & after & after \\
    includes & includes & includes \\
    is included & is included & is included \\
    simultaneous & simultaneous & simultaneous \\
    before & simultaneous & before \\
    after & simultaneous & after \\
    includes & simultaneous & includes \\
    is included & simultaneous & is included \\
    before & after & [before, after, includes, is included, simultaneous] \\
    before & includes & [before, includes] \\
    before & is included & [before, is included] \\
    after & before &  [before, after, includes, is included, simultaneous] \\
    after & includes & [after, includes] \\
    after & is included & [after, is included] \\
    includes & before & [before, includes] \\
    includes & after & [after, includes] \\
    includes & is included & [before, after, includes, is included, simultaneous] \\
    is included & before & [before, is included] \\
    is included & after & [after, is included] \\
    is included & includes & [before, after, includes, is included, simultaneous] \\
    simultaneous & before & before \\
    simultaneous & after & after \\
    simultaneous & includes & includes \\
    simultaneous & is included & is included \\ \hline
\end{tabular}
\caption{Transitivity rules for the five temporal relations used in this study.}
\label{tab:trans_relations}
\end{table*}

\begin{table*}[t!]
\centering
  \begin{tabular}{c c} \hline
    \bf $r$ & \bf $\bar{r}$ \\ \hline
    before & after \\
    after & before \\
    includes & includes \\
    is included & is included \\
    simultaneous & simultaneous \\ \hline
\end{tabular}
\caption{Symmetry rules for the five temporal relations used in this study.}
\label{tab:sym_relations}
\end{table*}

\end{document}